\documentclass{article}

\usepackage[final, nonatbib]{tackling_climate_workshop_style}

\usepackage[utf8]{inputenc} 
\usepackage[T1]{fontenc}    
\usepackage{hyperref}       
\usepackage{url}            
\usepackage{booktabs}       
\usepackage{amsmath,amsfonts}       
\usepackage{nicefrac}       
\usepackage{microtype}      
\usepackage{float, graphicx}
\usepackage{tabularx}
\usepackage{gensymb} 

\usepackage[belowskip=-1ex,aboveskip=1ex]{caption}
\usepackage[
        backend=biber,
        style=numeric,
        sorting=none,
        giveninits=true
    ]{biblatex}
\DeclareNameAlias{default}{family-given}
\title{Combining deep generative models with extreme value theory for synthetic hazard simulation: a multivariate and spatially coherent approach}

\author{%
  Alison Peard and Jim W.~Hall \\
  School of Geography and the Environment\\
  University of Oxford\\
  Oxford, UK \\
  \texttt{alison.peard@ouce.ox.ac.uk} \\
}

\begin{document}
\maketitle
\begin{abstract}
Climate hazards can cause major disasters when they occur simultaneously as compound hazards. To understand the distribution of climate risk and inform adaptation policies, scientists need to simulate a large number of physically realistic and spatially coherent events. Current methods are limited by computational constraints and the probabilistic spatial distribution of compound events is not given sufficient attention. The bottleneck in current approaches lies in modelling the dependence structure between variables, as inference on parametric models suffers from the curse of dimensionality. Generative adversarial networks (GANs) are well-suited to such a problem due to their ability to implicitly learn the distribution of data in high-dimensional settings. We employ a GAN to model the dependence structure for daily maximum wind speed, significant wave height, and total precipitation over the Bay of Bengal, combining this with traditional extreme value theory for controlled extrapolation of the tails. Once trained, the model can be used to efficiently generate thousands of realistic compound hazard events, which can inform climate risk assessments for climate adaptation and disaster preparedness. The method developed is flexible and transferable to other multivariate and spatial climate datasets.
\end{abstract}

\section{Introduction}
Compound events, defined as combinations of climate variables, not all of which are necessarily extreme, but which lead to extreme impacts \parencite{leonard2014compound}, can wreak far greater damages than those composed of a single driver \parencite{zscheischler2018future}. Most major recent climate catastrophes, such as droughts caused by low precipitation and high temperatures; and coastal flooding caused by co-occurring high tides, heavy precipitation, and strong onshore winds, were compound events. Univariate risk assessments have been shown to significantly underestimate true hazard levels \parencite{zscheischler2018future}. Additionally, hazard maps which assume spatially-homogeneous return periods are physically unrealistic and lead to inaccurate risk assessments \parencite{vorogushyn2018evolutionary}. For successful climate adaptation policies to be developed it is paramount that scientists develop accurate and efficient methods to model the spatial distributions of compound hazard events.

There are two classical approaches for modelling the distribution of multivariate extreme events: copulas \parencite{nelsonintroduction, davison2012statistical} and the multivariate conditional exceedence model \parencite{heffernan2004conditional}. Both approaches decompose the problem into modelling the univariate marginals and the dependence structure between them. Here, marginal refers to the probability distribution of a single hazard at a point in space. Copulas are based on Sklar's theorem \parencite{sklar1959fonctions} and accurate modelling relies on choosing the most appropriate copula family. The multivariate conditional exceedance model replaces the copula with a semiparametric conditional model that learns the conditional distribution of all the marginals, given that one of them is extreme, and introduces flexibility into the asymptotic dependence structure between pairs of variables. Both methods rely on parameter estimation, hence inference becomes computationally inefficient as dimensionality increases.

Machine learning is increasingly being used to address the shortcomings of traditional approaches in multivariate extreme modelling. \Textcite{letizia2020segmented} developed a fully nonparametric model using two separate GANs to capture both the marginal distributions and the dependence structures. \Textcite{bhatia2021exgan} developed ExGAN, which generates samples of desired extremeness levels for rainfall data over the United States using a conditional GAN. \textcite{boulaguiem2022modeling} combined extreme value theory with deep convolutional GANs (DCGANs) to generate annual maximum precipitation and temperature fields over Europe, training a separate, single-channel model for each of the two variables.

We develop an extension on the work of \textcite{boulaguiem2022modeling} to the multi-dimensional case by simultaneously training a DCGAN on three channels of gridded climate data: wind speed, significant wave height, and total precipitation. By training on daily maxima rather than annual maxima, we capture the propensity for extremes to co-occur on a given day. We present preliminary results and demonstrate that the model generates physically realistic samples that capture the dependence structure across space and between variables. Such a model can be used to efficiently generate a large synthetic ensemble of multivariate events, addressing the urgent need for better modelling of compound events in climate risk assessment \parencite{zscheischler2018future}.

\section{Methodology}
\paragraph{Data} Hourly wind speeds {[ms\textsuperscript{-1}]}, significant height of combined wind waves and swell {[m]}, and total precipitation {[m]} for the years 2013-2022 are obtained from the \textit{ERA5 hourly data on single levels from 1940 to present} reanalysis product \parencite{era5}.
The data has 0.25\degree\ spatial resolution. We obtain the data over the Bay of Bengal (longitude: 10-25\degree\ East,  latitude: 80-95\degree\ North) and calculate the daily maxima. We resize the images to $18 \times 22$ pixels using bilinear interpolation and zero-pad them to $20 \times 24$ pixels. 

\paragraph{Normalisation} It is conventional to normalise data before training a neural network, to stabilise gradient descent and improve convergence. \Textcite{boulaguiem2022modeling} propose to replace standard re-scaling methods with the probability integral transform \parencite{fisher1932statistical, casella2001statistical}, where each observation $X^j\in\mathbb{R}$ is replaced by its empirical distribution function $\hat F_X(X^j)$, which follows a uniform distribution over the interval [0,1]. This effectively separates the dependence structure from the marginal distributions. The neural network is then trained on the uniform-transformed data.  For the inverse transform,  \textcite{boulaguiem2022modeling}  fit a generalised extreme value (GEV) to each sequence of annual maxima in its original scale. The GEV parameters are shape $\xi$, location $\mu$, and scale $\sigma$.
\footnote{This fitting relies on assumptions of independence and stationarity that underlie the Fisher-Tippett-Gnedenko theorem \parencite{fisher1928limiting, gnedenko1943distribution, coles2001introduction} and the climate variables in this paper exhibit some autocorrelation and seasonality, as is typical for climate data.  However, it has been shown that the dependence assumption can be relaxed in cases where the long-range dependence at extreme levels is weak, as the data will still fall into the same family of GEV distributions (\cite{coles2001introduction} Chapter 5). Following a similar argument to \textcite{heffernan2004conditional}, we do not attempt to correct the seasonal component of the data as we are interested in the overall joint distribution of daily maxima, and not their temporal evolution.}
The shape parameter determines the shape of the tails, which are bounded below for $\xi > 0$ and bounded above for $\xi < 0 $ \parencite{coles2001introduction}. The corresponding probability point function is then used to transform generated uniform samples back to the original scale.

\begin{figure}[h!]
\centering
     \includegraphics[width=\textwidth]{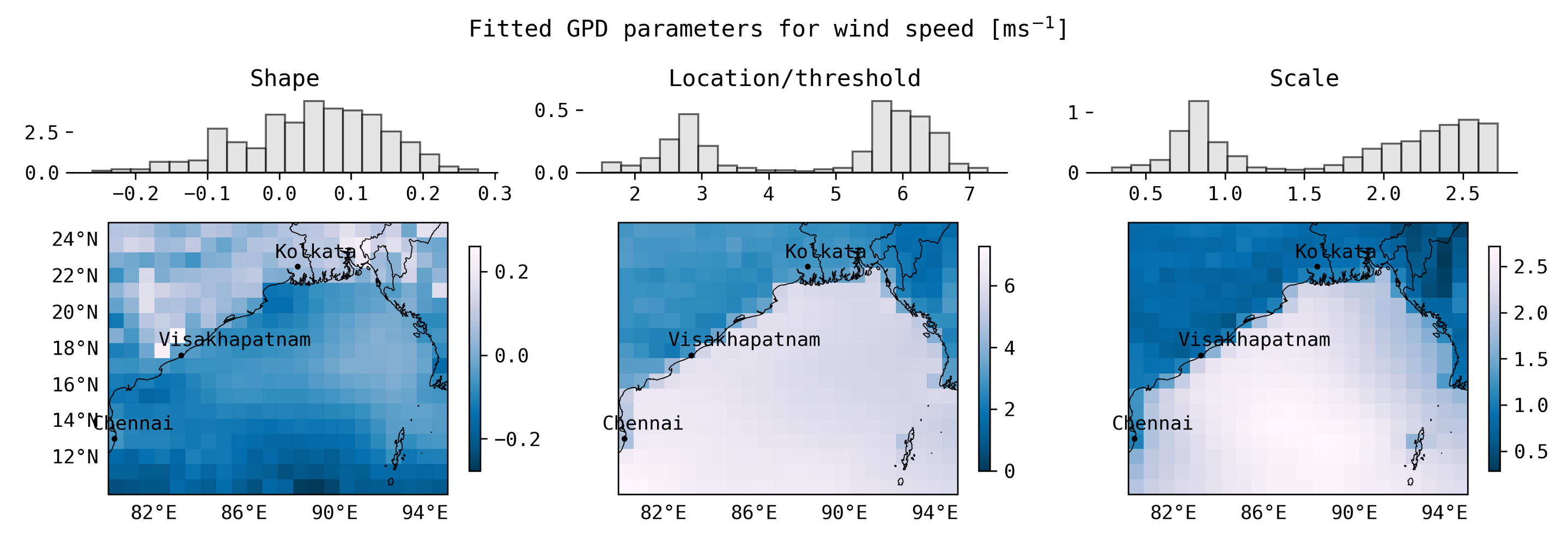}
     \caption{Fitted generalised extreme value (GEV) distribution parameters shape ($\xi$), location ($\mu$), and scale ($\sigma$), over the Bay of Bengal for wind speed in ms$^{-1}$.}
     \label{fig:fig1}
\end{figure}

Figure \ref{fig:fig1} shows the fitted parameters for wind speed over the Bay of Bengal (corresponding plots for significant wave height and precipitation are shown in Appendix \ref{appendix:tables_and_figures}). The fitted parameters indicate that wind is stronger and has higher variance offshore, but is also more likely to have a negative shape parameter $\xi$, indicating that it is bounded above. Describing offshore winds using a Weibull distribution is common practice, though more complex models have been proposed \parencite{morgan2011probability}. Significant wave height shows a relatively homogeneous distribution, with primarily positive $\xi$ indicating light-tailed distributions with no upper bound. Precipitation appears heavier and more varied over deep water and along the coastlines. The distribution of the shape parameter for precipitation appears less physical than other results. This may be due to the large number of zeros in the data biasing the fit. For the purposes of this paper, we leave further exploration of suitable marginal distributions as an opportunity for future work.

\paragraph{Deep convolutional GAN}
GANs \parencite{goodfellow2014generative}, are generative deep learning models composed of two competing neural networks which can implicitly learn the distribution of the training data \parencite{creswell2018generative} and capture local and global spatial patterns. The generator
$$G:\mathbb{R}^{100 \times M} \to \mathbb{R}^{20 \times 24 \times M},$$
with $M \in \mathbb{N}$ representing the number of samples, generates $M$ fake images of size $20 \times 24$ from a latent space $z \in \mathbb{R}^{100 \times M}$ with elements $ z_{i}^j \sim \text{Norm}(0, 1)$, and the discriminator
$$D:\mathbb{R}^{20 \times 24 \times M} \to [0, 1]^M,$$
attempts to distinguish fake data $G(\mathbf{Z})$ from real data $\mathbf{U}$. Elements of both $G(\mathbf{Z})$ and $\mathbf{U}$ have uniform distributions with support $[0, 1]$. The discriminator seeks to maximise the number of instances of $D(\mathbf{U}) = \mathbf{1}$ and minimise the instances of $D(G(\mathbf{Z}))=\mathbf{1}$. Likewise, the generator seeks to maximise the occurrences of $D(G(\mathbf{Z}))=\mathbf{1}$. This setup is captured in the cross-entropy objective function,
\begin{equation}
    \min_G \max_D \mathbb{E}_{\mathbf{U}}\left[\log D(\mathbf{U})\right] - \mathbb{E}_{G(\mathbf{Z})}\left[\log D(G(\mathbf{Z}))\right].
    \label{eq:gan_objective}
\end{equation}
Deep convolutional GANs (DCGANs) \parencite{radford2015unsupervised} employ convolutional and deconvolution layers, which can learn local patterns and features of image data in the generator and the discriminator, respectively. We train a DCGAN on the data transformed to uniform marginals, where each sample is an image consisting of three channels corresponding to wind speed, significant wave height, and total precipitation. We outline the architecture and training procedure in more detail in Appendix \ref{appendix:theory}.

\section{Results and discussion}
\begin{figure}[h!]
\centering
     \includegraphics[width=\textwidth]{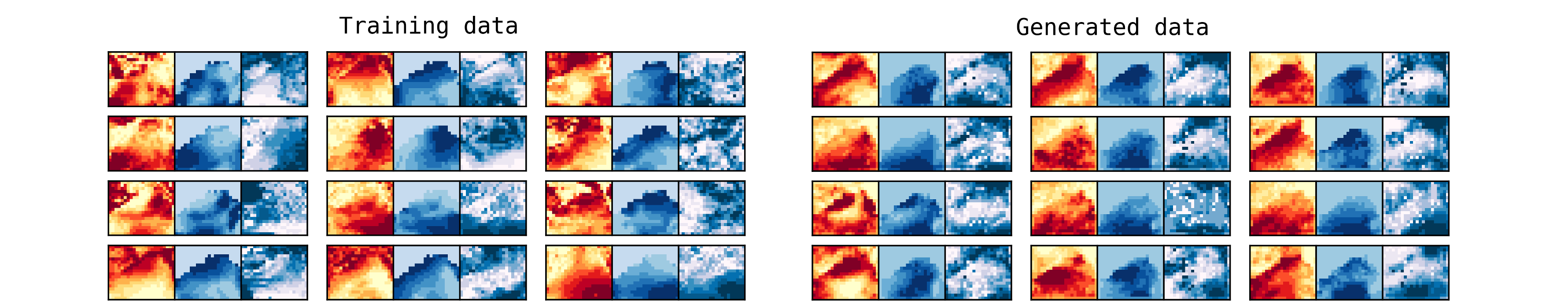}
     \caption{Heatmaps for the training (left) and generated (right) data. Daily maximum wind speed, significant wave height, and precipitation are shown side-by-side for each sample, where darker colours indicate higher values in each case. All heatmaps of the same variable have the same colour scale. (Larger figures for both are shown in Appendix \ref{appendix:tables_and_figures}.)}
     \label{fig:fig2}
\end{figure}

We present results for data generated by a DCGAN model trained for 1000 epochs over a training set of size $N=1000$ with a batch size of 50, then transformed to the original scale using the fitted GEV distributions. Figure \ref{fig:fig2} shows a sample from the training (left block) and generated (right block) data, where each group of three images represents daily maximum fields for wind speed (left), significant wave height (centre), and total precipitation (right). (Larger figures are shown in Appendix \ref{appendix:tables_and_figures}.) Visually, the generated images appear similar to the training images, with good variation between samples. Wind and waves display higher values in similar locations on the same day in both sets. Significant wave height accounts for wind-sea waves, which are driven by local winds \cite{era5}, and it's clear the GAN has learned this relationship.

\begin{figure}[h!]
\centering
     \includegraphics[width=.8\textwidth]{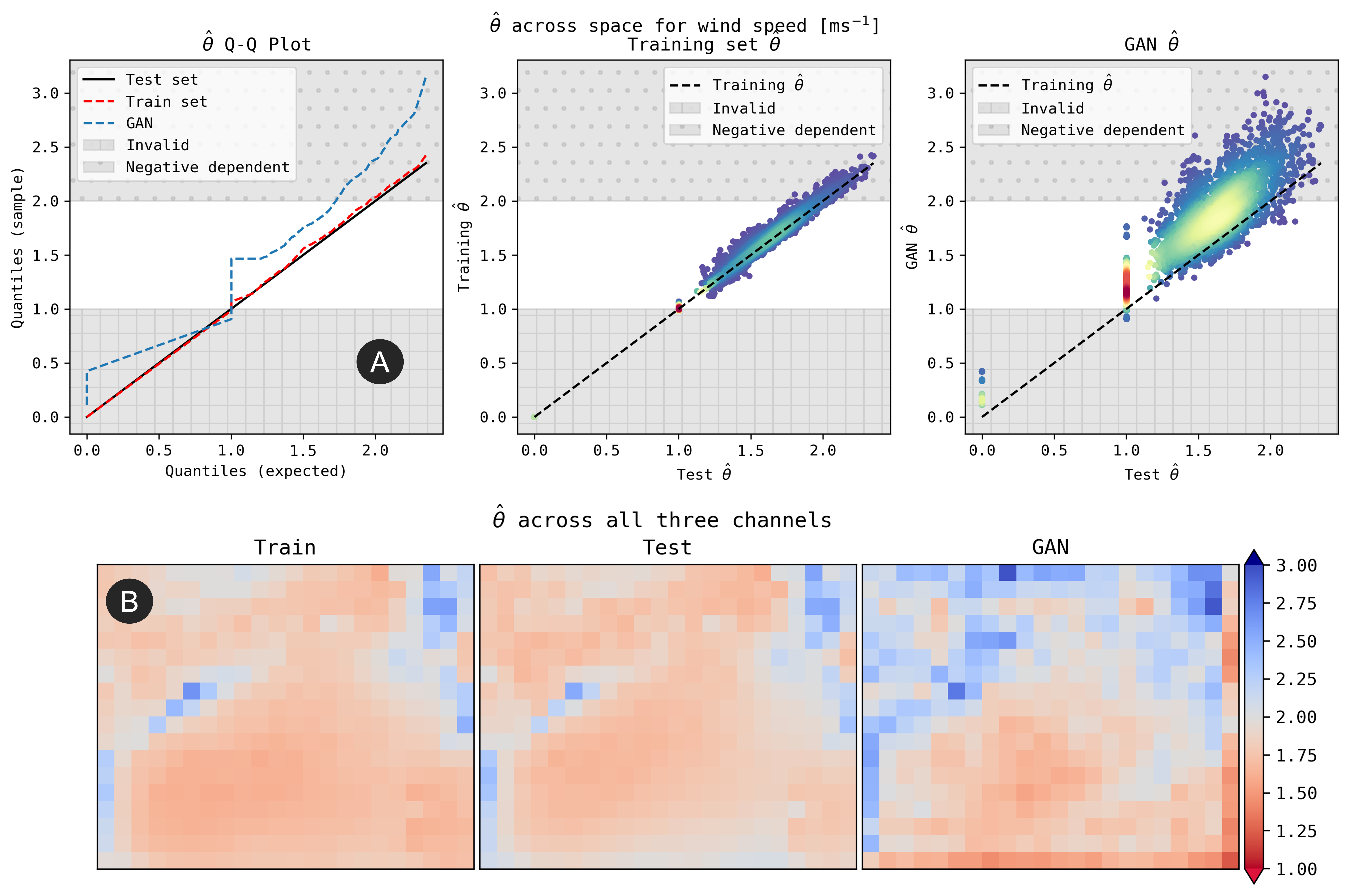}
         \caption{\textbf{A:} Distributions of $\hat \theta$ across space for wind data for all $4950$ possible pixel pairs: Q-Q plots comparing test, train, and generated sets (left), and scatter plots comparing the train and generated set to the test set (centre and right). \textbf{B:} Heatmaps showing three dimensional extremal coefficient estimates $\hat \theta$ for the train, test, and generated data. The GAN-generated data shows more noise, and a bias towards underestimation, particularly near land, as seen by the higher $\hat \theta$ in all plots, but has learned the general spatial structure of the multivariate extremal dependencies.}
     \label{fig:fig3}
\end{figure}

To quantify the extremal dependency between variables, we use the extremal coefficient $\theta$ \parencite{davison2012statistical, smith1990max, strokorb2012characterizing}, which takes values in the range $[1, D]$ for a $D$-dimensional sample, indicating total dependence or independence between variables, respectively. In a three-dimensional setting with $N$ samples, this can estimated using,

$$\hat \theta_{123} = \frac{N}{\sum_{n=1}^N\min\left(\frac{1}{Y_{n1}}, \frac{1}{Y_n2}, \frac{1}{Y_{n3}}\right)}$$
where the $Y_{ni}$ have unit Fréchet distributions, which are easily obtained from the uniformly distributed marginals using $Y_{ni} = -\log(U_{ni})^{-1}$. The extremal coefficient can be understood as the effective number of independent variables in the sample, while the extremal correlation $\hat \chi = 1- \hat \theta$ forms an extreme analogue for the Pearson correlation coefficient. Comprehensive discussions and derivations of these coefficients are given in \textcite{davison2012statistical} and \textcite{smith1990max}. Figure \ref{fig:fig3} compares the distribution of the estimator $\hat \theta$ between the train, test, and generated sets across {[A]} space and {[B]} the wind, wave, and precipitation dimensions. Additonal figures in Appendix \ref{appendix:tables_and_figures} show the model has learned the distribution of the spatial correlations and the spatial extremal correlations.

Overall, the model has learned the extremal dependence structure across space and climate variables but exhibits bias towards larger $\hat \theta$, i.e., underestimating the dependence between the most extreme events. Analysis of the generated data indicates the DCGAN is generating fewer samples near the highest percentiles for each marginal. GANs generally produce samples from the mean of the training distribution so this is not wholly unexpected \parencite{bhatia2021exgan, huster2021pareto} and may be improved by replacing the standard normal latent space vectors with a more heavy-tailed distribution \parencite{huster2021pareto} or by transfer learning on extreme subsets of the training data.

\section{Conclusion}
In this paper, we have demonstrated the usefulness of deep generative models for learning dependence structures in high dimensional climate data, and how this can be combined with extreme value theory to facilitate interpretability and to control extrapolation to extreme values for spatial, compound events. We have applied this methodology to a dataset of daily maximum wind speed, significant wave height, and precipitation from 2013 to 2022 over the Bay of Bengal and developed a model that can efficiently generate a large number of synthetic compound events.

The model learns the dependence structure well both across space and between different variables, though it displays more overall noise and has a tendency to underestimate the most extreme events. This underestimation is likely due to GAN's tendency to reproduce samples from the centre of a distribution and could potentially be improved by using heavier-tailed distributions, such as the Student's $t$-distribution, for the latent space variables. 
To avoid potential side-effects, we stress that when using this model as part of a larger project, care should be taken to ensure the input data is reliable, or that its limitations are explicitly known, as the model will propagate any errors in the input data into generated data.

There are ample opportunities for future development of this work, including a deeper exploration of distributions to be fitted to the univariate marginals; the use of alternative deep generative models such as diffusion models, which have more stable training, and normalising flow models, which explicitly learn the data distribution; or by making improvements to the current model through architecture modifications, further parameter tuning, and transfer learning on extreme subsets of the training data.

The methods in this paper construct a computationally efficient model for generating synthetic datasets of compound events that are coherent in space and time. Such datasets can be used as inputs for further risk modelling and are critically important for understanding the true spatial distribution of the risk of compound extreme events and informing climate adaptation and coastal disaster preparedness policies.

\begin{ack}
The authors would like to thank Geoff Nicholls and Steven Reece for their valuable feedback on this work. This work was supported by the Engineering and Physical Sciences Research Council (EPSRC).
\end{ack}

\newpage
\printbibliography
\appendix
\section{Theory}\label{appendix:theory}
\subsection{Normalisation using extreme value theory}\label{appendix:pit}
\paragraph{Fisher-Tippett-Gndenko theorem}
Given a sequence of $n$ independent and identically-distributed (i.i.d.) random variables $X^1, X^2, ..., X^n$ with a common distribution function $\mathbb{P}(X^j\leq x)=F(x)$, the Fisher-Tippett-Gnedenko theorem \parencite{fisher1928limiting, gnedenko1943distribution} provides us with results about the behaviour of the sample maximum $Y^{(n)} = \max(X^1, X^2, ..., X^n)$ as $n$ approaches infinity, in a manner analogous to the central limit theorem for the sample mean \parencite{coles2001introduction}. It is easily shown that the distribution function of $Y^{(n)}$ is given by $[F(x)]^n$, however this becomes degenerate for large $n$. Hence a sequence of shift and scale parameters $\{b_n\}_{n\in \mathbb{N}}$ and $\{a_n: a_n > 0\}_{n\in \mathbb{N}}$, is applied. The limit is instead defined for the distribution of $$\tilde Y^{(n)} = \frac{Y^{(n)} - b_n}{a_n}.$$The Fisher-Tippett-Gnedenko theorem shows that the limiting distribution must be one of the three distributions which make up the generalised extreme value (GEV) distribution: Gumbel, Fréchet, or Weibull. The GEV is defined three parameters $(\xi, \mu, \sigma)$, corresponding to its shape, location, and scale, respectively.

In practice, however, the $X^j$ are rarely i.i.d. Climate data has trends, seasonality, and dependencies in time. This motivates a \textit{block maxima} approach to extreme value modelling, where the maxima are taken over intervals of size $k$ of the observations. For sufficiently large $k$, e.g., $k=365$ for annual maxima, the maxima can be assumed to be independent and approximately stationary, and a GEV is fitted to these block maxima. \Textcite{coles2001introduction} shows that, provided dependence becomes neglible for sufficiently far apart samples, the independence assumption can be relaxed as autocorrelation only affects the shape and scale parameters but not what the shape parameter.

\subsection{DCGAN architecture and training}\label{appendix:dcgan}
A training set of 2000 days is sampled from the data, and the rest is withheld as a test set. We refer to the sequence of daily maxima for a single spatial location and variable as a marginal. The entire training set is used to fit a GEV distribution to each marginal and subset of size $N$ of this set is then used for training the DCGAN.

The generator consists of an input layer that takes a 100-dimensional tensor $z\in\mathbb{R}^{100}$, followed by a dense layer with 25,600 neurons. The output of the dense layer is then reshaped into a $5\times 5\times 1024$ tensor and passed through batch normalization, a leaky ReLU activation function, and a dropout layer. The tensor is then passed through two transposed convolutional layers, each with 512 filters, followed by batch normalization, a leaky ReLU activation function, and a dropout layer. The final layer is a transposed convolutional layer with three filters and a sigmoid activation function. The total number of trainable parameters in the generator is 10,490,883.

\begin{figure}[H]
\centering
     \includegraphics[width=\textwidth]{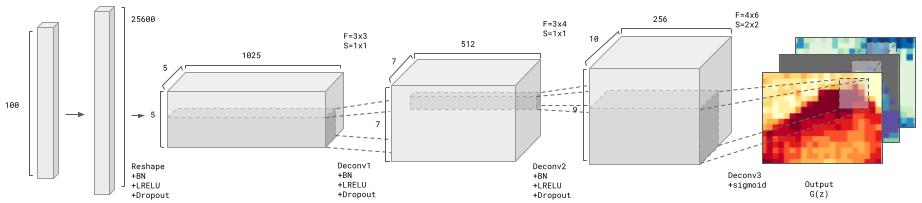}
     \caption{DCGAN generator architecture, adapted to the multivariate case from \textcite{boulaguiem2022modeling}. Layers are indicated below the blocks, LRELU stands for a leaky ReLU layer, BN for a batch normalisation layer, and sigmoid for a sigmoid activation layer. Above the blocks, F represents the filter size, and S the stride of the deconvolutional laters. The output has three channels and dimensions of $20 \times 24$.}
     \label{fig:generator_architecture}
\end{figure}

The discriminator has three convolutional layers with 64, 128, and 256 filters respectively, each followed by LeakyReLU activation and dropout. The output of the last convolutional layer is reshaped to $1 \times 6400$. The output is then reshaped to a scalar and passed through a sigmoid activation function to scale predictions to the [0, 1] interval, where $0$ labels the input data as fake and $1$ labels it as real. The model has a total of 405,057 parameters, with 404,289 of them being trainable.

\begin{figure}[H]
\centering
     \includegraphics[width=\textwidth]{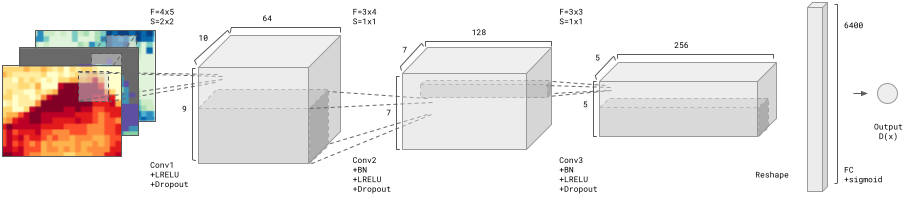}
     \caption{DCGAN discriminator architecture, adapted to the multivariate case \textcite{boulaguiem2022modeling} Layers are indicated below the blocks, LReLU stands for a leaky RELU layer, BN for a batch normalisation layer, FC for a fully-connected layer, and sigmoid for a sigmoid activation layer. Above the blocks, F represents the filter size, and S the stride of the deconvolutional laters. The output has is a scalar in the interval [0,1].}
     \label{fig:discriminator_architecture}
\end{figure}

GANs are notoriously unstable to train and several modifications have been suggested to improve their stability. We use label smoothing, where real images are assigned a label of 0.9 instead of 1, in the discriminator objective. Further modifications to the DCGAN architecture have been suggested, such as the Wasserstein GAN \parencite{arjovsky2017wasserstein} which replaces the cross-entropy objective with the first Wasserstein distance and places additional constraints on the discriminator. However, motivated by the results of \textcite{lucic2018gans} which find correct hyperparameter selection to be more powerful in improving GAN performance than architecture modifications, we focus on hyperparameter tuning rather than architecture changes.

We use the WandB \parencite{wandb} machine learning experiment tracking tool to perform a comprehensive Bayesian grid search and tune the hyperparameters. A table of all the hyperparameters and configurations explored, along with the final selections, is shown in Appendix \ref{appendix:tables_and_figures}.

\newpage
\section{Tables and figures}\label{appendix:tables_and_figures}
\subsection{Tables}
\begin{table}[H]
\caption{Parameters included in the WandB Bayesian grid search and final selections.}
\begin{center}
\renewcommand{\arraystretch}{1.5}
\begin{tabularx}{\textwidth}{XccX}
\hline
\textbf{Parameter} & \textbf{Values} & \textbf{Selected Value} & \textbf{Description}\\
\hline
\hline
seed & [0, 1, 2, 6, 7, 42] & 7 & Included in sweep for reproducibility.\\
learning\_rate & 0.0001 - 0.0003 & 0.00013367626823798716 & Controls the optimizer step size and rate of convergence. \\
beta\_1 & 0.1 - 0.5 & 0.22693882275467836 & Controls the exponential decay rate for the first moment estimates of the gradient for the Adam optimizer. \\
lrelu & 0.1 - 0.4 & 0.2991161912395133 & The gradient to assign to negative values in the Leaky ReLU function. \\
dropout & 0.3 - 0.6 & 0.44053850596844424 & Frequency the dropout function sets inputs to zero in training to prevent overfitting. \\
training\_balance & [1, 2] & 2 & Ratio of training loops for discriminator vs. generator. \\
\hline
\end{tabularx}
\label{table:wandb_sweep}
\end{center}
\label{table:dcgan}
\end{table}

\subsection{Figures}
\subsection*{GEV parameters}
\begin{figure}[h!]
\centering
     \includegraphics[width=\textwidth]{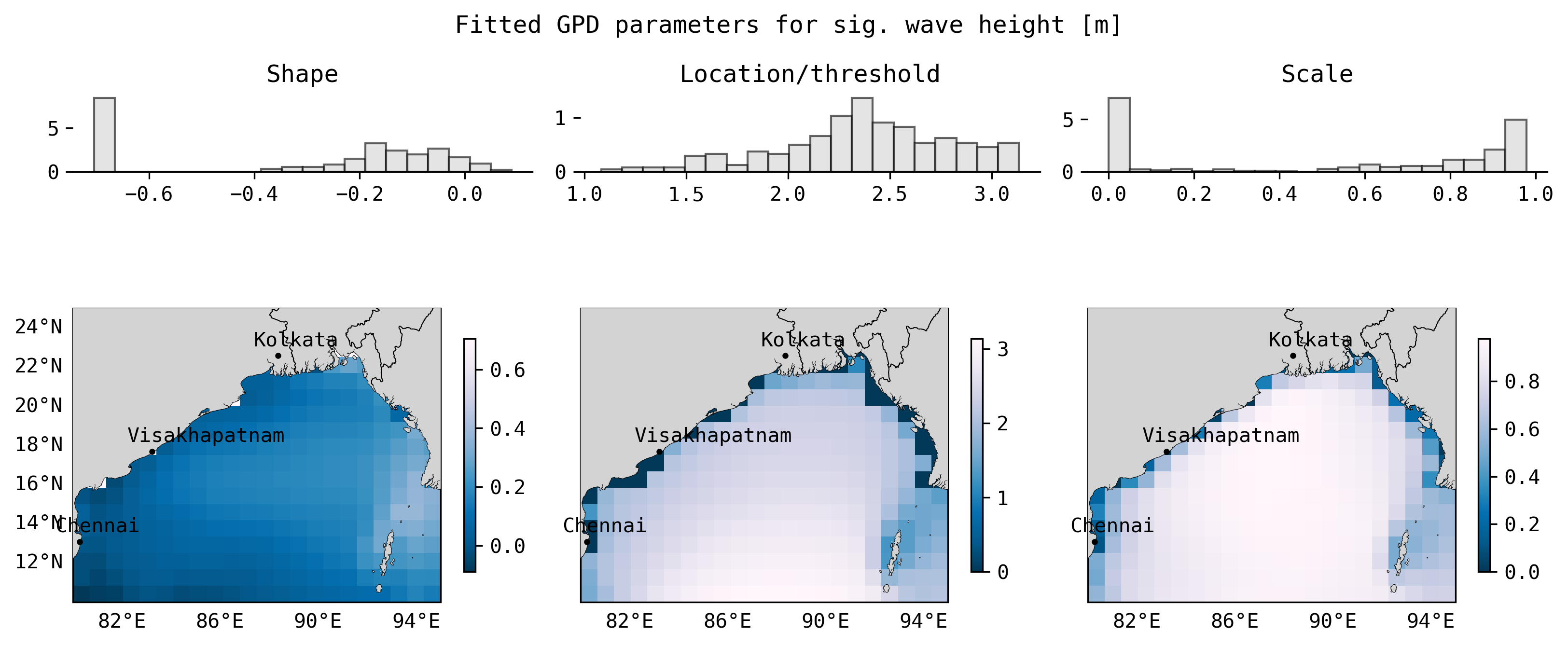}
     \caption{Fitted generalised extreme value (GEV) distribution parameters shape, location, and scale, over the Bay of Bengal for significant wave height from wind waves and swell in m. Here, the value -999 is used on land to represent no data.}
     \label{fig:params_wave_data}
\end{figure}
\begin{figure}[H]
\centering
     \includegraphics[width=\textwidth]{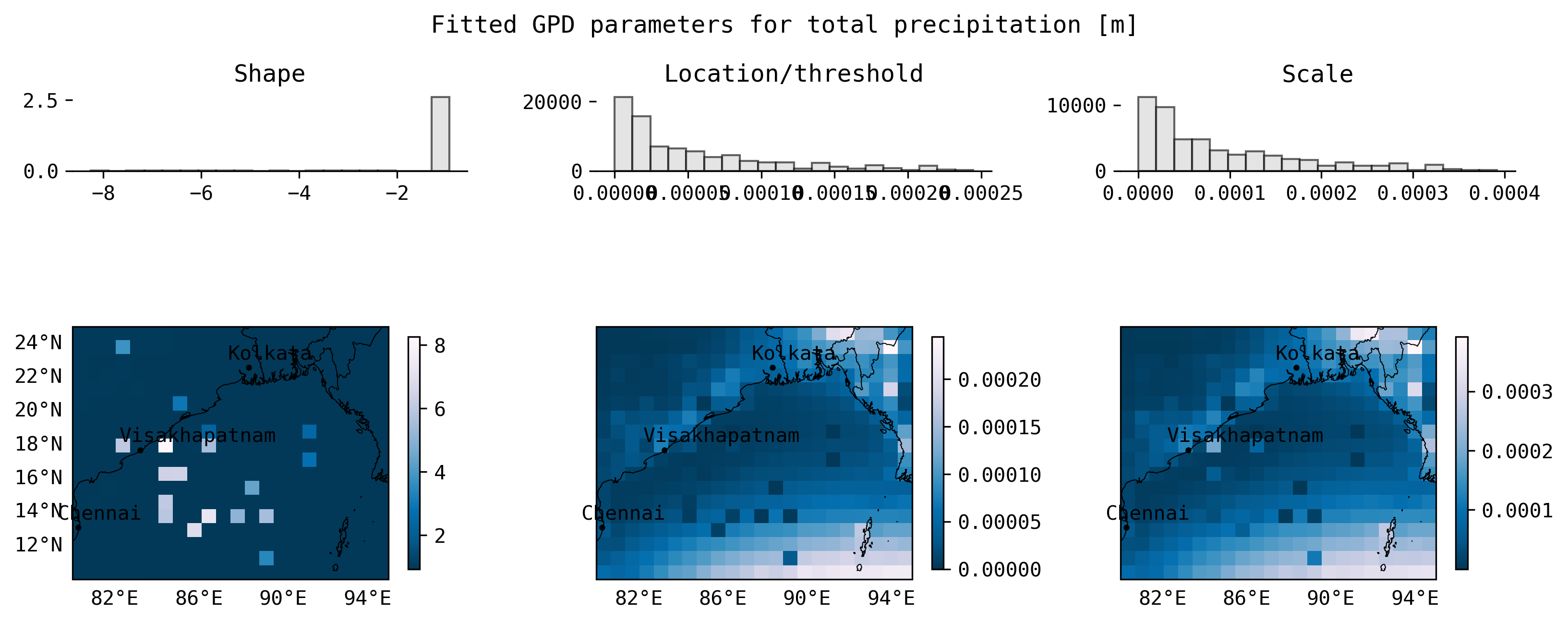}
     \caption{Fitted generalised extreme value (GEV) distribution parameters shape, location, and scale, over the Bay of Bengal for total precipitation m.}
     \label{fig:params_precip_data}
\end{figure}
\subsection*{Training and generated data}
\begin{figure}[H]
\centering
     \includegraphics[width=\textwidth]{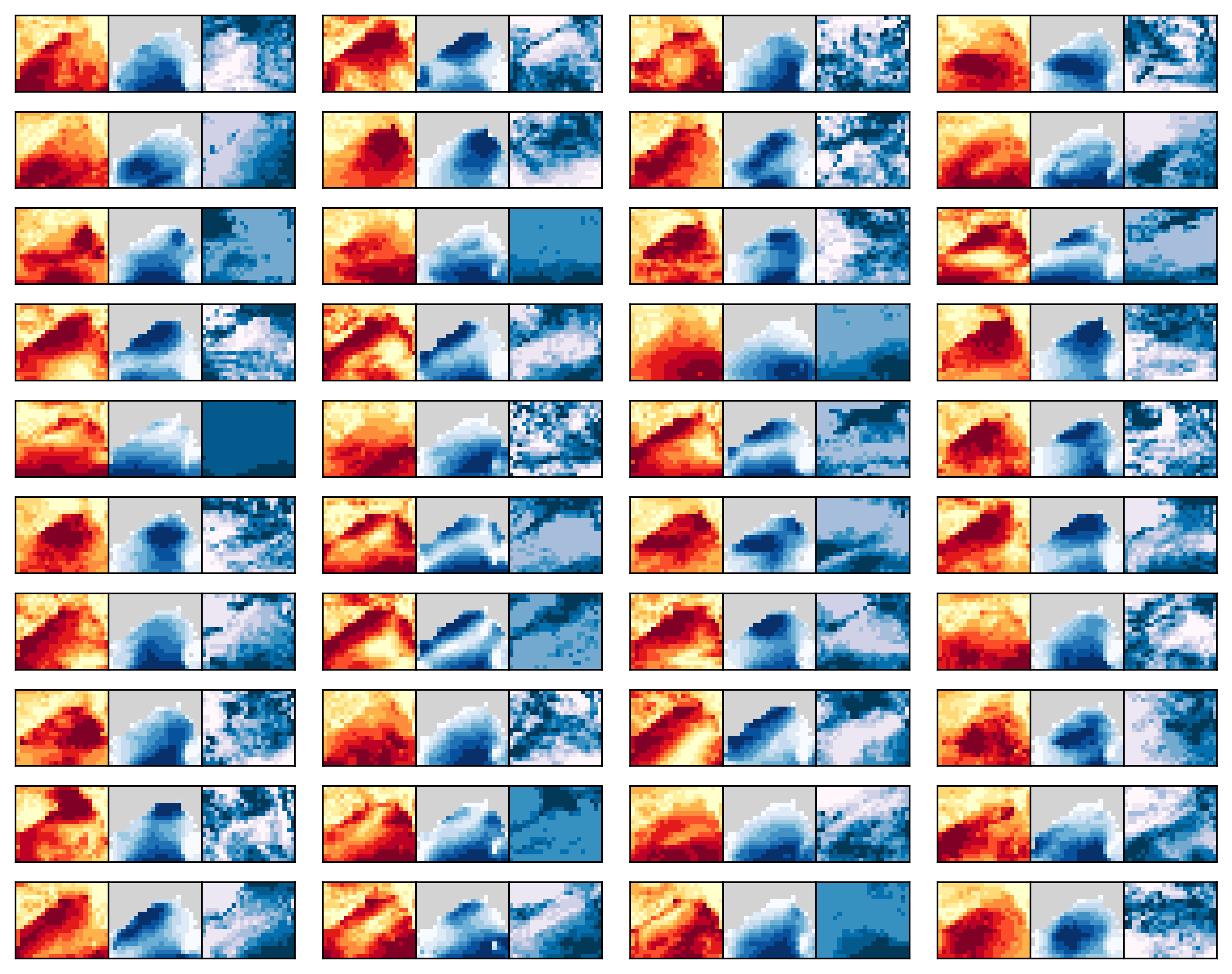}
     \caption{Sample of training data. Daily maximum wind speed (left), significant wave height (centre), and precipitation (right) are shown side-by-side for each sample.}
     \label{fig:train_data}
\end{figure}
\begin{figure}[H]
\centering
     \includegraphics[width=\textwidth]{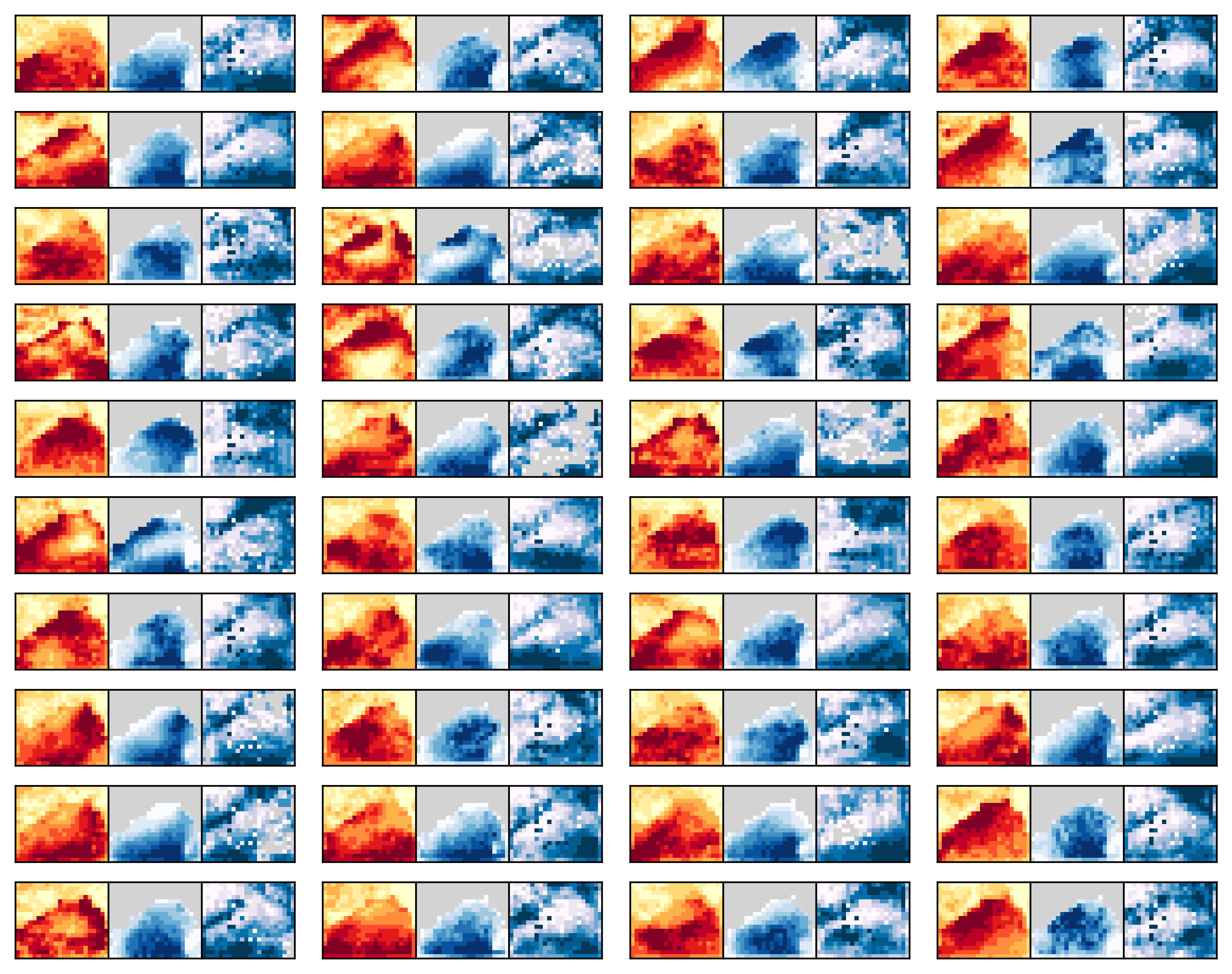}
     \caption{Sample of generated data. Daily maximum wind speed (left), significant wave height (centre), and precipitation (right) are shown side-by-side for each sample.}
     \label{fig:generated_data}
\end{figure}
\subsection*{Multivariate extremal coefficients}
\begin{figure}[H]
\centering
     \includegraphics[width=\textwidth]{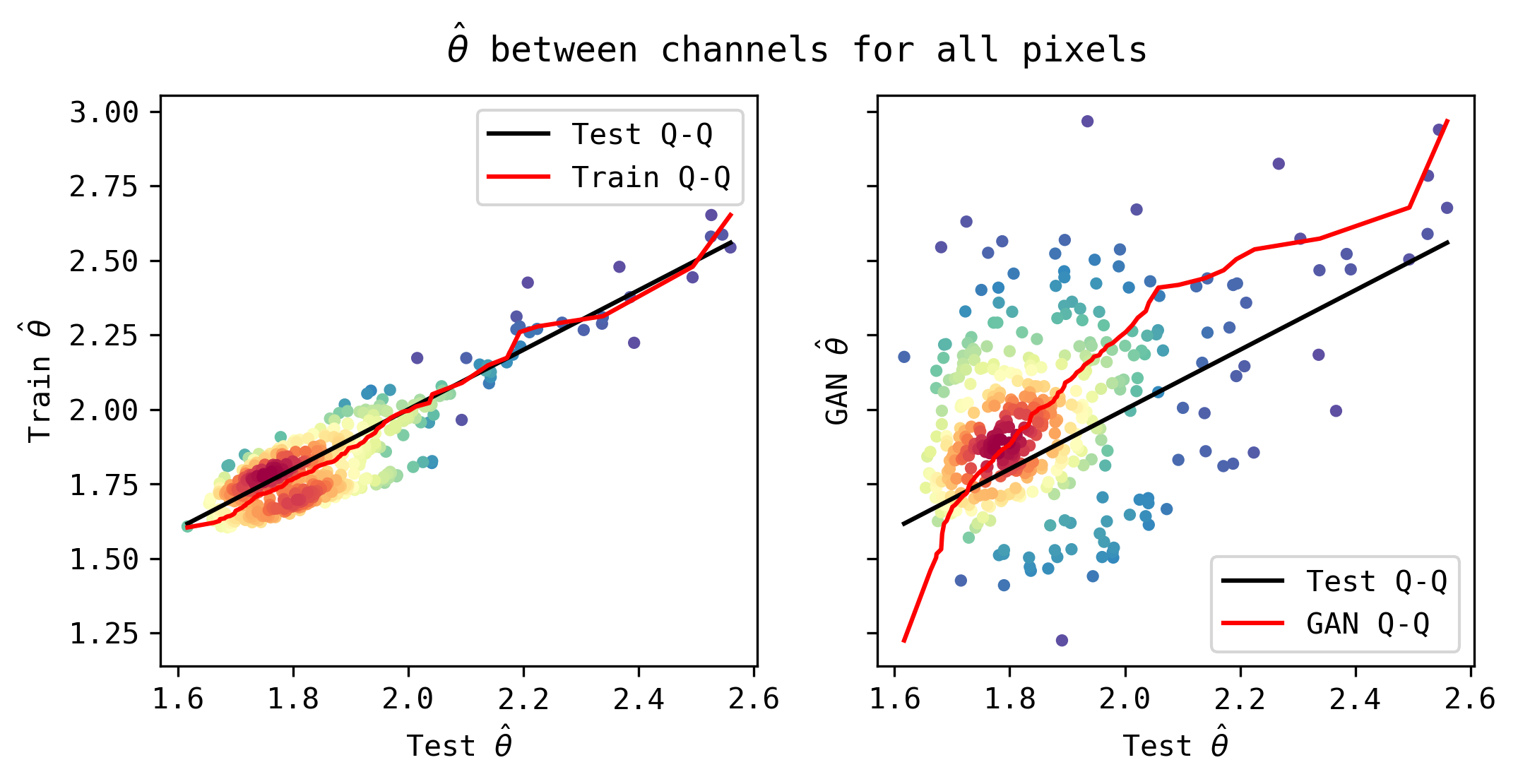}
     \caption{Scatter plots of extremal coefficient $\hat \theta$ across all three channels for all 396 pixels for the test and train sets (left) and test and generated sets (right). Solid red lines show the Q-Q plots for the train and generated sets versus the test set.}
     \label{fig:ecorr_multichannel_scatter}
\end{figure}
\subsection*{Spatial correlations}
\begin{figure}[H]
\centering
     \includegraphics[width=.8\textwidth]{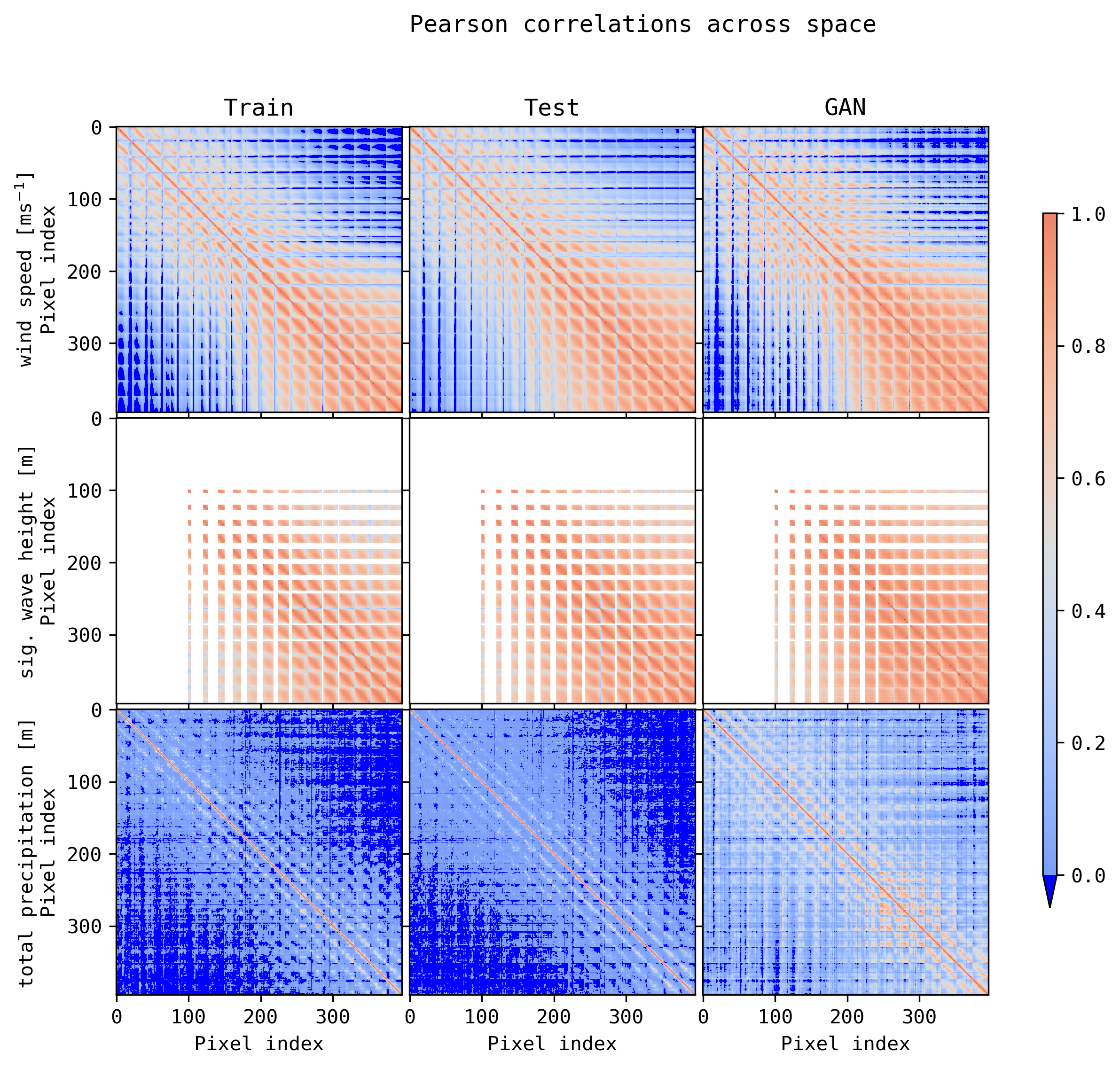}
     \caption{Matrix of pairwise Pearson correlations between all $18\times 22=396$ pixels for the training, test, and generated data. Clearly, the GAN captures the spatial correlation structure between the variables almost identically for wind speed (top row) and significant wave height (middle row), but underestimates the negative correlations for total precipitation (bottom row). We see a pattern of more positive correlations between nearby pixels for wind and precipitation. A group of positively correlated pixels emerges in the bottom right corner of the wind heatmaps. This corresponds to offshore pixels being more correlated to each other than to land pixels and aligns with observations made about the GEV fit to the wind data.}
     \label{fig:spatial_corrs}
\end{figure}
\begin{figure}[H]
\centering
     \includegraphics[width=.8\textwidth]{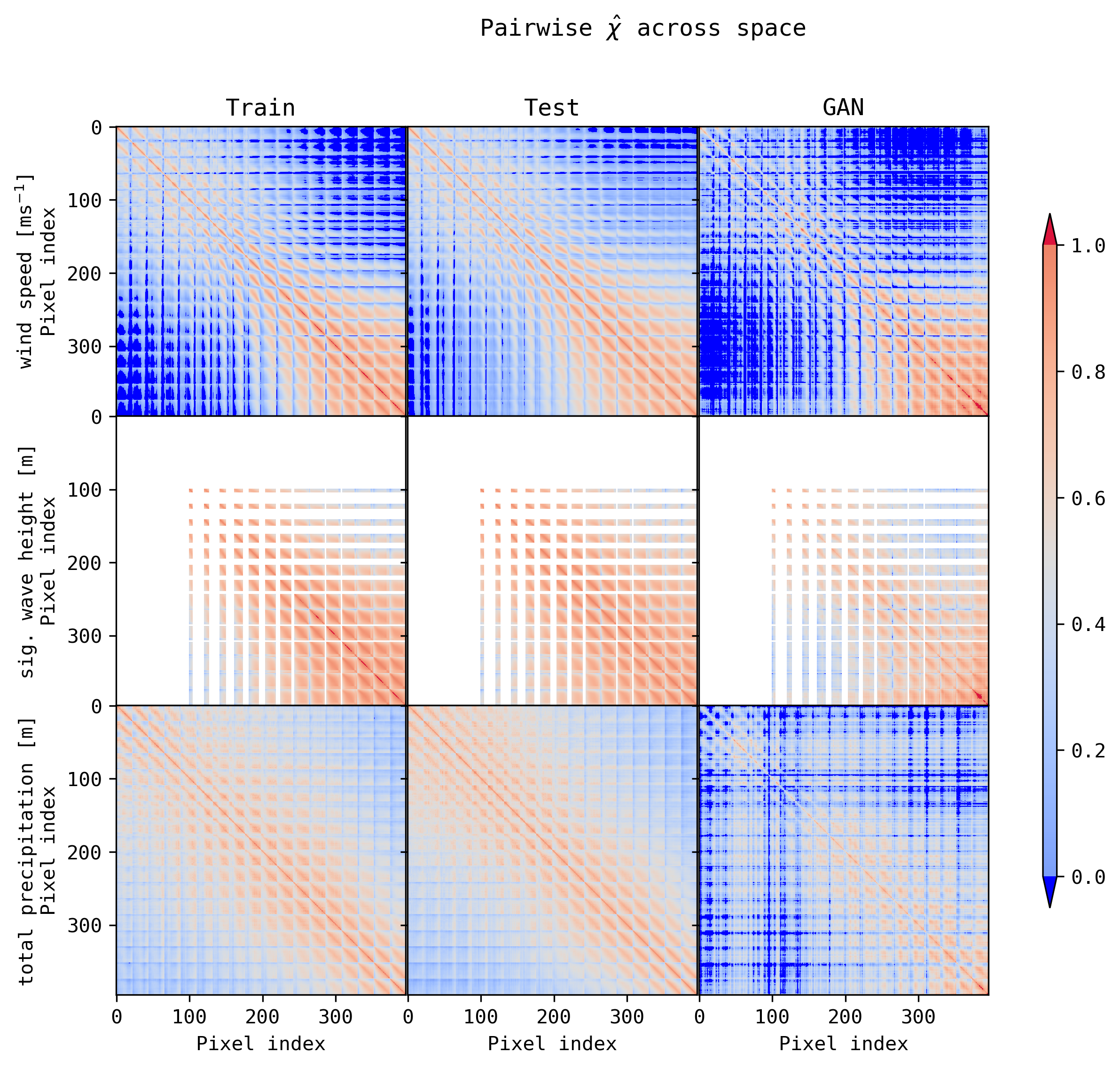}
     \caption{Matrix of pairwise extremal correlations $\hat \chi$ between all $18\times 22=396$ pixels for the training, test, and generated data. The GAN captures the overall spatial pattern of the extremal correlations, correctly identifying that extreme winds (top row) offshore are more positively correlated to each other than to onshore extreme winds. The significant wave height (middle row) extremal correlations follow the spatial distribution train and test sets, with a tendency to underestimate extremal correlation as distance increases. Total precipitation (bottom row) appears to capture the overall distribution but appears noisier, with more negative $\hat \chi$s corresponding to the land pixels in the top left (approx pixels 1 to 150) of the heatmap.}
     \label{fig:spatial_corrs}
\end{figure}
\end{document}